\documentclass[twocolumn,10pt]{article}
\usepackage{amsmath,amssymb,amsfonts}
\usepackage[utf8]{inputenc}
\usepackage{graphicx}
\usepackage{textcomp}
\usepackage{xcolor}
\usepackage{cite}
\usepackage[utf8]{inputenc}
\usepackage{graphicx}
\usepackage{xcolor}
\usepackage{bbm}
\usepackage{bm}

\usepackage{abstract} 
\usepackage{authblk} 

\title{RSI-Grad-CAM:
  Visual Explanations from Deep Networks via Riemann-Stieltjes Integrated
  Gradient-based Localization}

\author[1,3]{Mirtha Lucas}
\author[2]{Miguel Lerma}
\author[1]{Jacob Furst}
\author[1]{Daniela Raicu}
\affil[1]{DePaul University, Chicago, IL 60604, USA}
\affil[2]{Northwestern University, Evanston, IL 60208, USA}
\affil[3]{\textit{mlucas3@depaul.edu}}

\begin{document}
\twocolumn[
\maketitle
\begin{onecolabstract}
Neural networks are becoming increasingly better at tasks that involve 
classifying and recognizing images.
At the same time techniques intended to explain the network output have been proposed.
One such technique is the Gradient-based Class Activation Map (Grad-CAM), which is able to locate
features of an input image at various levels of a convolutional neural network (CNN), but 
is sensitive to the vanishing gradients problem. There are techniques such as Integrated Gradients (IG), 
that are not affected by that problem, but its use is limited to the input layer of a network.
Here we introduce a new technique to produce visual explanations for
  the predictions of a CNN. Like Grad-CAM, our method can be applied to any layer of the
  network, and like Integrated Gradients 
    it is not affected by the problem of
    vanishing gradients.  For efficiency, gradient
  integration is performed numerically at the layer level
  using a Riemann-Stieltjes sum approximation.  Compared to Grad-CAM,
  heatmaps produced by our algorithm are better focused in the areas of interest,
    and their numerical computation is more stable.
    Our code is available at \texttt{https://github.com/mlerma54/RSIGradCAM}.
\end{onecolabstract}
]

\section{Introduction}

The visualization of features captured by convolutional neural
networks (CNN) helps explain how they make their predictions.  This
is a field of rapid development in which many techniques have been
proposed, tested, and validated; methods to provide
explanations for the predictions of a CNN can be grouped into three
main categories: primary attribution methods, layer attribution methods and neuron attribution methods \cite{kokhlikyan2020captum}.  

\emph{Primary
  attribution} methods evaluate the contribution of each input to the
output of a model.  
This approach is model-agnostic, meaning that primary attribution methods work
the same regardless of the internal structure of the network or
machine learning system used, in that depends only on inputs and outputs, 
not on internal structure. 
Some examples are Integrated Gradients (IG) \cite{sundararajan2017ig} and
Local Interpretable Model-Agnostic Explanations
(LIME)~\cite{ribeiro2016why}. 

\emph{Layer attribution} methods evaluate the
contribution of each neuron in a given layer to the output of the
model.   These methods are useful to determine the location of 
medium and high level features such as the spacial location of the various elements
that compose an image.  Some examples are
 Class Activation Mapping (CAM) \cite{zhou2016cam} and its derivatives such as 
 Gradient-based CAM (Grad-CAM) \cite{selvaraju2017grad} 
and Grad-CAM++~\cite{chattopadhyay2018}.

\emph{Neuron attribution} methods evaluate the contribution of
each input feature to the activation of given hidden neurons. Both primary and layer attribution methods can be converted
into neuron attribution methods by replacing the network output with 
any neuron activation.  For instance Neuron Integrated Gradients
is equivalent to Integrated Gradients considering the output to be simply
the output of the identified neuron.  They may be useful when we are
interested in determining how the activation of a given neuron, 
rather than the network output, depends on the input of the network.
It is also possible to combine the effect of the input on a given neuron and the 
effect of the neuron on the network output, as in the Neuron Conductance 
method described in \cite{dhamdhere2018important}.

In our study we look at three gradient based methods: 
Gradient Guided Class Activation Map (Grad-CAM, a layer
attribution method) \cite{selvaraju2017grad}, 
Integrated Gradients (a primary attribution method)
\cite{sundararajan2017ig}, and Integrated Grad-CAM (layer attribution) \cite{sattarzadeh2021igcam}.
We examine their advantages and limitations, and propose a
modification of Grad-CAM in which gradients are replaced with
integration of gradients computed at any layer rather than the
input layer. 
This allows our method to simultaneously overcome the limitations of grad-CAM, and Integrated Gradients,
namely vulnerability to the vanishing gradients problem, and applicability to the network input only
respectively.

\section{Previous Work}

Our attribution method combines ideas from two existing techniques:
Grad-CAM and Integrated Gradients. 
In this section we explain how these methods work.
We also look at the Integrated Grad-CAM technique
introduced in \cite{sattarzadeh2021igcam}, and 
show how it differs from ours.

\subsection{Grad-CAM}

The technique introduced in \cite{selvaraju2017grad} uses the
gradients of any target concept flowing into a convolutional layer to
produce a heatmap, also called saliency map 
or localization map,\footnote{We will be using the terms \emph{heatmap}, 
\emph{saliency map}, and \emph{localization map} interchangeably.}
intended to highlight the contribution of
regions in the image for predicting the concept.

\begin{figure}[!htb]
\centering
\includegraphics[width=3.4in]{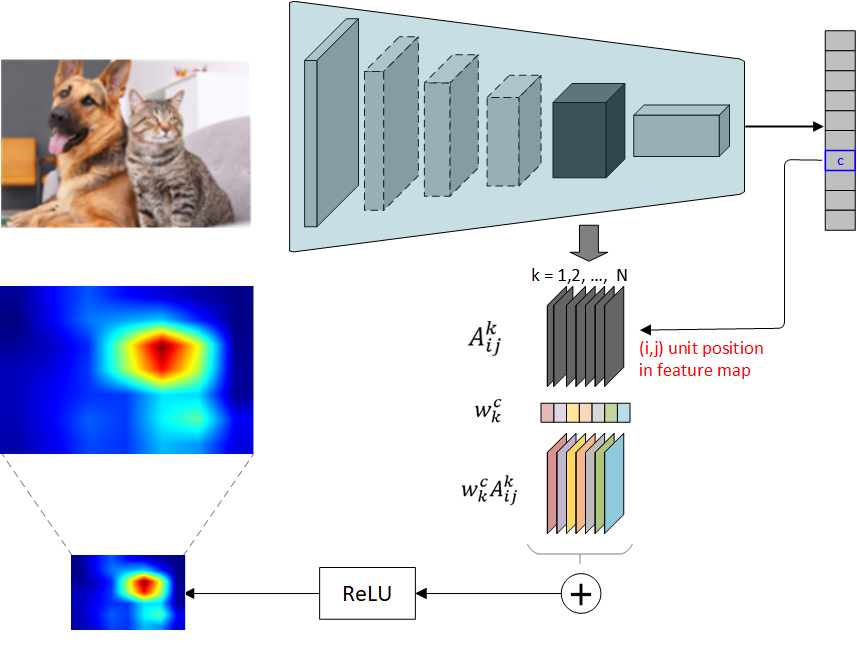}
\caption{Grad-CAM overview}\label{f:gradcam}
\end{figure}

Grad-CAM works as follows (see {\figurename}\,\ref{f:gradcam}).  First we must pick a convolutional layer
$A$, which is composed of a number of feature maps, also called
channels, $A^1, A^2, \dots, A^{N}$ (where $N$ is the number of feature maps in the picked layer), all of them with the same 
dimensions. At the network input they are just the three RGB channels of the input image.
In the following layers the channels/feature maps are supposed to capture progressively
higher features of the image (beginning with 
edge detection, shapes, textures, geometric shapes, 
and ultimately whole categories such as ``dog'' and ``cat'').
We will use the term ``channels'' when referring to the third dimension
of a layer, and ``feature maps'' if we want to stress their feature capturing
role.

Let $A^k$ be the $k$-th feature map of the picked
layer, and let $A_{ij}^k$ be the activation of the unit in the
position $(i,j)$ of the $k$-th feature map.
Then, the localization map,
or ``heatmap,'' is obtained by combining the feature maps of the layer
using weights $w_k^c$ that capture the contribution of the $k$-th
feature map to the output $y^c$ of the network corresponding to class~$c$.

In order to compute the weights, we pick a class~$c$ and determine how
much the output $y^c$ of the network depends of each unit of the
$k$-th feature map, as measured by the gradient
$\partial y^c/\partial A_{ij}^k$, which can be obtained by
using the backpropagation algorithm.  The gradients are then averaged
thorough the feature map to yield a weight $w_k^c$, as indicated in
equation (\ref{e:gradcam_weights}). Here $Z$ is the size (number of
units) of the feature map.

\begin{equation}\label{e:gradcam_weights}
  w_k^c = \overbrace{\frac{1}{Z} \sum_{i}\sum_{j}}^{\text{global average pooling}}
  \hskip -30pt
  \underbrace{\frac{\partial y^c}{\partial A_{ij}^k}}_{\text{gradients via backprop}}
\end{equation}

Optionally, the weights could be computed using positive gradients:
\begin{equation}\label{e:gradcam_weights2}
  w_k^c = \frac{1}{Z} \sum_{i}\sum_{j}
\text{ReLU}\left(\frac{\partial y^c}{\partial A_{ij}^k}\right)
\end{equation}
Where $\text{ReLU}(x) = \text{max}(x,0)$ is the Rectified Linear Unit
function. This can be justified by the
intuition that negative gradients correspond to units where
features from a class different from the chosen class are present.

The next step consists of combining the feature maps $A^k$ with the
weights computed above, as shown in equation (\ref{e:heatmap}).  Note
that the combination is also followed by a Rectified Linear Unit function
$\text{ReLU}(x) = \text{max}(x,0)$,
because we are interested only in the features that have a positive
influence on the class of interest. The result $L_{\text{Grad-CAM}}^c$
is called \emph{class-discriminative localization map} by the authors. 
It can be interpreted as a coarse heatmap of the same size as the chosen
convolutional feature map.

\begin{equation}\label{e:heatmap}
  L_{\text{Grad-CAM}}^c =
  \text{ReLU} \underbrace{\Biggl(\sum_k w_k^c A^k\Biggr)}_{\text{linear combination}}
\end{equation}

\begin{figure}[!htb]
\centering
\includegraphics[width=3.4in]{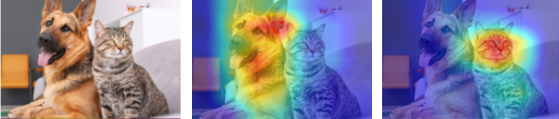}
\caption{Original image (left). Grad-CAM locating a dog (center) and a cat (right).}\label{f:catdog}
\end{figure}

After the heatmap has been produced, it can be normalized and
upsampled via bilinear interpolation to the size of the original image,
and overlapped with it to highlight the areas of the input image that
contribute to the network output corresponding to the chosen class (see {\figurename}\,\ref{f:catdog}).

The method is very general, and can be applied to any (differentiable)
network outputs.

In spite of its success, Grad-CAM has a limitation. It has the
possibility of getting a very small or practically zero gradient 
($\partial y^c/\partial A_{ij}^k \approx 0)$ 
when the network output is near saturation, i.e., when the value 
of the output is very close to its maximum (say $100\%$ score assigned to a class). In 
this instance, any increase in the value is very small (almost zero).
This \emph{vanishing gradient problem} was first noticed in the
context of training neural networks by the backpropagation algorithm 
(see e.g. \cite{hochreite1998}), but it can appear in any application based on
backpropagating gradients.  If the saturation occurs at the network
output, then it can be palliated by
replacing the final output of the network (typically a softmax) with
the activations of the layer right before the final softmax (as in the
description included in \cite{selvaraju2017grad}), however the problem
can potentially affect any layer.

\subsection{Integrated Gradients}

A related attribution technique, introduced in
\cite{sundararajan2017ig}, is Integrated Gradients~(IG).  The authors study
the problem of finding a method to attribute predictions of a deep
network to its input features that verifies two desirable properties
(called ``axioms'' by the authors).  To formulate them we need a
baseline input in which all features are absent (e.g. a blank image),
and a given input in which some features are present (say the image of
a fireboat). The properties are the following:

\begin{itemize}

\item Sensitivity: For every input and baseline that differ in one
  feature but have different predictions, the differing feature should
  be given a non-zero attribution.

\item Implementation invariance: Two functionally equivalent networks
  should have identical attributions for the same input image and 
  baseline image.

\end{itemize}

\begin{figure}[!htb]
\centering
\includegraphics[width=2.5in]{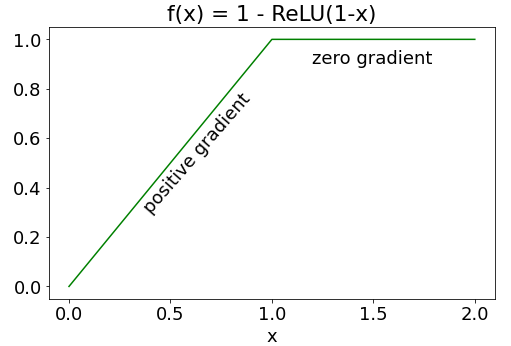}
\caption{Example of violation of the sensitivity property
because of vanishing gradient.}\label{f:one_relu_net}
\end{figure}

The sensitivity property is violated by Grad-CAM and other methods
relying on gradients. This happens because a relevant feature may be given no
attribution due to vanishing gradient when the network output gets close to
its saturation point. The authors present this example: consider a
one variable, one ReLU network implementing the following function: 
\begin{equation}\label{e:relu_network}
f(x) = 1 - \text{ReLU}(1-x)
\end{equation}
Its graph is shown in {\figurename}\,\ref{f:one_relu_net}.

Suppose the input is $x=2$ and its associated baseline is $x=0$.
Then, the function changes from
$f(0) = 0$ to $f(2) = 1$, but $f(x)$
becomes flat for $x>1$, and the gradient method gives attribution
$f'(2) = 0$. But this contradicts the fact that the value of the
function has changed, so the attribution should not be zero.

The method proposed in \cite{sundararajan2017ig}
is immune to that problem because it does not depend only on one
gradient at a given level of activation, but on the result of
integrating gradients along a set of network inputs obtained by
interpolating between a \emph{baseline} input (e.g. a black
image) and the actual desired input.

The authors describe their method with a theoretical model that uses
continuous functions, and later describe how to implement it in a
discrete setting.  They interpret a network as a multivariate
function $F:\mathbb{R}^d\to[0,1]$ from its $d$ inputs to the
prediction $F(x)$ of the network for a given input
$x\in \mathbb{R}^d$.  In case $x$ represents an image, then $d$ would be its
number of pixels, and $F(x)$ may represent the probability that the
image contains a given object, say a fireboat.  The goal would be to
determine which pixels in the image contribute to the prediction of
the network, in other words, which of those pixels are part of the
fireboat image.

Hence, instead of using a single image the method uses a
sequence of interpolated images between a baseline $x'$ and the given
image $x$:
\begin{equation}\label{e:inter_imag}
\gamma(\alpha) = x' + \alpha (x-x') \qquad 0 \leq \alpha \leq 1
\end{equation}

Each interpolated image is a combination  
of $\gamma(0) = x'$ (baseline) and $\gamma(1) = x$ (given image).
Then, the gradient of the network output 
with respect to each input pixel $x_i$
is integrated as shown in equation (\ref{e:igint}).

\begin{multline}\label{e:igint}
  \texttt{IntegratedGrads}_i(x) ::= \\
  (x_i - x'_i) \times \int_{\alpha=0}^{1}
  \frac{\partial F(x' + \alpha \times (x - x'))}{\partial x_i} \, d\alpha
\end{multline}

In the equation, the factor 
$(x-x')$ appears when using as variables of
integration the pixel values of the interpolated image, so that
the differential within the integral is
$d(x' + \alpha \times (x - x')) = (x-x') \times d\alpha$.
Since it does not depend on $\alpha$, the factor $(x-x')$ can be taken 
outside the integral.

In practice the integral can be approximated numerically with a Riemann sum:
\begin{multline}\label{e:igapprox}
 \texttt{IntegratedGrads}_i^{approx}(x) ::= \\
 (x_i - x'_i) \sum_{\ell=1}^m
 \frac{\partial F(x' + \frac{\ell}{m}\times (x-x'))}{\partial x_i} \times \frac{1}{m}
\end{multline}
where $m$ represents the number of interpolation steps. This parameter
$m$ can be adjusted by experimentation, although it is recommended to give it
a value between 50 and 200.\footnote{The most common pixel format is the 
byte image, where a number is stored as an 8-bit integer, giving 
a range of possible values from 0 to 255. An interpolating process with 200 steps
is likely to exhaust all intermediate possible values of most pixels
between baseline and final image.}

\begin{figure}[!htb]
\centering
\includegraphics[height=3.4in]{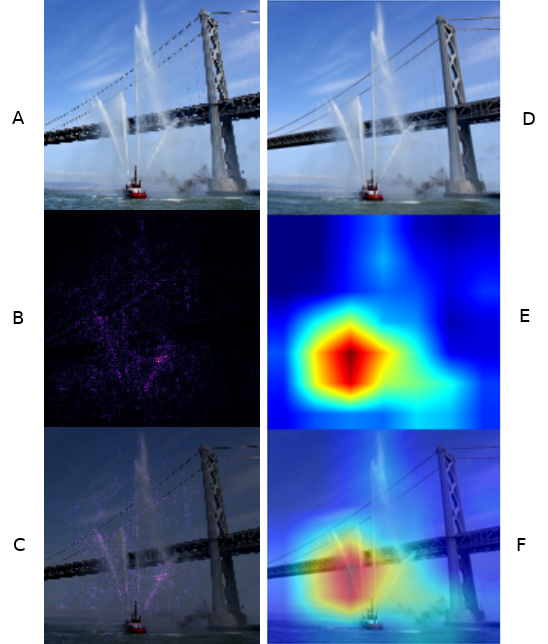}
\caption{A fireboat image located by Integrated Gradients (left) and Grad-CAM (right).
For integrated gradients it is shown:
(A) input image, (B) IG attribution mask, (C)
overlay of attribution mask with original image. 
For Grad-CAM we show (D) the original image,
(E) heatmap, (F) overlay.}\label{f:fireboat}
\end{figure}

The result plays the role of a localization map, although compared to
Grad-CAM, Integrated Gradients tends to look more grainy, 
while Grad-CAM heatmaps are
smoother (see {\figurename}\,\ref{f:fireboat} as an example).

The IG attribution mask is the output of Integrated Gradients for this
figure, showing the contribution of each input pixel to the class ``fireboat.''

A problem with the Integrated Gradients method is that it is designed
for working with the network inputs and may miss features captured
at hidden layers. A workaround consists of considering the activations
of a hidden layer $A$ as input of the part of the network above that layer,
as suggested in the \emph{layer integrated gradients} method listed in 
\cite{kokhlikyan2020captum}. 
The role of the inputs now will be played by the 
activations of layer~$A$, and the value of $F$ will be the output of the 
network $y^c$ for some fix class $c$.
Let $A_{\text{baseline}}$ be the activations of layer $A$
when we feed
the network with the baseline input $x'$, and let $A_{\text{final}}$ 
be the activations of layer $A$ when the we feed
the network with the final input $x$. Then a linear interpolation
at the layer level yields
\begin{equation}\label{e:ig_at_layer_inter}
A(\alpha) = 
A_{\text{baseline}} + \alpha \, (A_{\text{final}} - A_{\text{baseline}})
\qquad 0 \leq \alpha \leq 1
\end{equation}
Let $y^c(\alpha)$ be the network output for a value of $\alpha$ between 0~and~1.
Using equation (\ref{e:igapprox}) we get a localization map $L$ with the same
shape as $A$ and elements
\begin{equation}\label{e:ig_to_layer}
L_{ij}^k =
 (A_{ij}^k(1) - A_{ij}^k(0))  \left( \frac{1}{m} \sum_{\ell=1}^m
 \frac{\partial y^c(\alpha_{\ell})}{\partial A_{ij}^k} \right)
\end{equation}
The result can be expressed in a more compact way as follows:
\begin{equation}\label{e:ig_to_layer_compact}
L = (A(1) - A(0)) \odot  \left( \frac{1}{m} \sum_{\ell=1}^m
 \frac{\partial y^c(\alpha_{\ell})}{\partial A} \right)
\end{equation}
where $\odot$ represents the Hadamard (element-wise) product \cite{horn2012}.
For visualization purposes $L$ can be summed across its channels,
i.e., $L' = \sum_{k=1}^{N} L^k$, max-min normalized, and finally resized
to the size of the original image.

We note that the axiomatic properties of IG applied
to a hidden layer may still hold for the part of the network above
the chosen layer, but it won't hold for the whole network---in
particular the method won't be implementation invariant anymore.
Another issue concerns the choice of baseline for a hidden layer.
While such choice can be guided by heuristic arguments when dealing
with network inputs such as images,
it is less clear when dealing with latent features at
the layer level---what set of activations can be taken to mean ``absence''
of a feature?

\subsection{Integrated Grad-CAM}\label{igradcam}

Suitable combinations of the ideas behind Grad-CAM and Integrated
Gradients can lead to new techniques that overcome the shortcomings
of each of them, particularly:

\begin{itemize}
\item Violation of the sensitivity axiom by Grad-CAM.
\item Applicability of Integrated Gradients to network inputs only.
\end{itemize}

Next, we will discuss the Integrated Grad-CAM technique introduced in
\cite{sattarzadeh2021igcam}. 

Combining equations (\ref{e:gradcam_weights}) and (\ref{e:heatmap}) we get the
following expression for the final heatmap in the original Grad-CAM:
\begin{equation}\label{e:gradcam}
L_{\text{Grad-CAM}}^c = \text{ReLU} \Biggl(\sum_k  \frac{1}{Z} \sum_{i,j}
  \frac{\partial y^c}{\partial A_{ij}^k} A^k \Biggr) 
\end{equation}

Similarly to Integrated Gradients, the authors use a set of
interpolated images between a baseline $x'$ and a final image $x$
as shown in (\ref{e:inter_imag}).
Those
images are fed to the network, and for each of them the activations of
the chosen layer and the gradients of the score $y^c$ respect to the
activations $A_{ij}^k$ are computed.  Then, an \emph{explanation map} $M^c$
(playing the role of Grad-CAM's $L^c$ localization map)
is generated using the following formula:
\begin{equation}\label{e:igradcam}
M^c =  \int_{\alpha=0}^{1} \text{ReLU} \Biggl(\sum_k  \sum_{i,j}
  \frac{\partial y^c(\alpha)}{\partial A_{ij}^k}
  \Delta_k(\alpha)\Biggr) \, d\alpha
\end{equation}
where $\Delta_k(\alpha) = A^k(\alpha) - A^k(0)$.
Here
$\partial y^c(\alpha)/\partial A_{ij}^k$ 
represents the partial derivative of
$y^c$ with respect to $A_{ij}^k$ when the input 
$\gamma(\alpha) = x' + \alpha (x-x')$ is
fed to the network, and $A_{ij}^k(\alpha)$ is the value of the
activation in location $(i,j)$ of the $k$-th feature map of the chosen
layer.  The explanation map can be computed numerically
using a Riemann sum for the integral:
\begin{equation}\label{e:igradcam_approx}
M^c \approx \sum_{\ell=1}^m  \text{ReLU} \Biggl(\frac{1}{m} \sum_k  \sum_{i,j}
  \frac{\partial y^c(\alpha_{\ell})}{\partial A_{ij}^k}
  \Delta_k(\alpha_{\ell})\Biggr)
\end{equation}
where $m$ is the number of interpolation steps, and $\alpha_{\ell} = \ell/m$.
Finally, $M^c$ is upsampled to the dimensions of the input image
via bilinear interpolation.

Comparing (\ref{e:gradcam}) and (\ref{e:igradcam}) we see that the
Integrated Grad-CAM equations differs from Grad-CAM in the following:

\begin{itemize}
\item The factor $1/Z$ is missing, which is not really
  relevant since the heatmap obtained is typically normalized to
  a fix interval of intensities, hence
  a global constant factor can be ignored.
\item The activations of the feature maps $A^k$ are replaced with
  the differences $\Delta_k(\alpha) = A^k(\alpha) - A^k(0)$
  between activations obtained when the network is fed with an
  interpolated image and the ones corresponding to the baseline.
\item The final explanation map is integrated along the set of
  interpolated images.
\end{itemize}

In sum, the method is equivalent to averaging Grad-CAM saliency maps
obtained for multiple copies of the input, which are linearly
interpolated with the defined baseline.

\section{Methodology}

In this section we will introduce a novel attribution method
combining ideas from Grad-CAM and Integrated Gradients, but essentially
different from the approach used in Integrated Grad-CAM.  
Also, our method, unlike Integrated Gradients, is not based on
axioms.  The authors of IG express skepticism about empirical
evaluations of attribution methods, for that reason their work is
based on axioms capturing desirable properties of an attribution method.
Our work will rely on empirical evaluations.
In the next section we will provide metrics showing
how our method overperforms Grad-CAM and Integrated Grad-CAM.

Like the Integrated Gradients attribution method, our algorithm feeds the network
with a set of inputs obtained by interpolation between a baseline and
a final input.  Then, it computes feature-map weights to be used like
in Grad-CAM, except that instead of gradients it uses the integral of
those gradients to compute the weight assigned to each feature-map.

The computation of the integrated gradients is formally equivalent to
the numerical computation of a Riemann-Stieltjes Integral \cite{protter1991integral}.
We start with a brief explanation of the concepts and mathematical
techniques, and then show how to use them to produce the three
attribution methods: Grad-CAM, Integrated Grad-CAM, and our RSI-Grad-CAM.
We finish with a summary comparing the methods discussed.

\subsection{Motivation and Theoretical Background}
Our technique aims to replace the gradients of the activations
$A_{ij}^k$ 
used by Grad-CAM with their integral as the network is fed
by a sequence of interpolated images (recall that $k$ indexes the 
feature maps within a given layer, and $(i,j)$ is the location of each
of the units of the feature map).  The main motivation is that we
are not interested in how much the output network $y^c$ for a given
class $c$ changes for an infinitesimal change of the activations
$A_{ij}^k$, but how much it changes along the whole interval of
values taken by each activation $A_{ij}^k$
as the image fed to the network 
goes from baseline to final image.
The idea
behind this technique is inspired by
the \emph{gradient theorem for line integrals}
\cite[p.\,374]{williamson2004mult}: a line integral through a gradient field
$\nabla F$ of a scalar vector field $F\!:\!\mathbb{R}^n \to \mathbb{R}$
along a given curve $\gamma\!:\![0,1]\to\mathbb{R}^n$ equals the
difference between the values of the scalar field at the endpoints
$\mathbf{p} = \gamma(0)$ and $\mathbf{q} = \gamma(1)$ of the curve:
\begin{multline}\label{e:gradthm}
  F(\mathbf{q}) - F(\mathbf{p})  = \\
  \int_{\gamma} \nabla F(\mathbf{x}) \cdot d\mathbf{r} = 
  \int_{\gamma} \sum_{i=1}^n \frac{ \partial F}{\partial x_i} dx_i 
  \sum_{i=1}^n \int_{\gamma} \frac{ \partial F}{\partial x_i} dx_i
\end{multline}
Each term $\int_{\gamma} \frac{ \partial F}{\partial x_i} dx_i$ of 
the final sum is the contribution of the $i$-th variable $x_i$ to
the total change of~$F$.
The gradient theorem is the basis of the
completeness property of Integrated Gradients
\cite[proposition~1]{sundararajan2017ig}.
In our case, the function will be the output of the network for a given
class $y^c$, and the variables of integration will be the activations
$A_{ij}^k$. The gradients are~$\partial y^c/\partial A_{ij}^k$,
and the term corresponding to the contribution of unit $(i,j)$ in 
feature map $k$ will be
$\int_{\gamma} \frac{\partial y^c}{\partial A_{ij}^k} \, dA_{ij}^k$.

Note that the $A_{ij}^k$ are not independent variables, but
(potentially complicated) functions of the network inputs.  An
integral in which the variable of integration is replaced with a
function is called a Riemann-Stieltjes integral \cite{protter1991integral}.
In general the integral of a function $f$ with respect to another
function $g$ is expressed like this:
\begin{equation}\label{e:stieltjes}
\int_a^b f(x) \, dg(x)
\end{equation}
where $g(x)$ is called the \emph{integrator}. In our problem the
activations $A_{ij}^k$ will play the role of
integrators.
This kind of integral can be numerically approximated
with a modification of a Riemann sum as follows:
\begin{equation}\label{e:ssum}
\int_a^b f(x) \, dg(x) \approx \sum_{\ell=1}^{m} f(x_{\ell}) [g(x_{\ell}) - g(x_{\ell-1})]
\end{equation}
where $x_{\ell} = a + \tfrac{\ell}{m} (b-a)$.

\subsection{Riemann-Stieltjes Sum Approximation}

Here the idea is to use a Riemann-Stieltjes integral like
(\ref{e:stieltjes}), with
$f = \partial y^c/\partial A_{ij}^k$ as integrand, and
$g = A_{ij}^k$ in the role of integrator, 
so the weight assigned to feature-map $k$ of
the chosen layer for a given class $c$ will be:
\begin{equation}\label{e:igcam_sint}
  w_k^c = \frac{1}{Z}
  \sum_{i,j} \int_{\alpha=0}^{\alpha=1} \frac{\partial y^c(\alpha)}{\partial A_{ij}^k} \, d A_{ij}^k(\alpha)
\end{equation}
where $\alpha$ is the interpolating parameter varying between $0$ and~$1$.

Note that (\ref{e:igcam_sint}) is just a line integral of the gradient
of $y^c$ as a function of the activations $A_{ij}^k$, along the
parametric curve $A^k(\alpha)$ joining $A^k(0)$ and $A^k(1)$.

The approximate value of the integral in (\ref{e:igcam_sint}) is given by the
following \emph{Riemann-Stieltjes sum}, as in (\ref{e:ssum}):
\begin{multline}\label{e:rs-sum}
\int_{\alpha=0}^{\alpha=1} \frac{\partial y^c(\alpha)}{\partial A_{ij}^k} \, d A_{ij}^k(\alpha)
\approx \\
\sum_{\ell=1}^m
    \Biggl\{\frac{\partial y^c(\alpha_{\ell})}{\partial A_{ij}^k}
    \times \Delta A_{ij}^k(\alpha_{\ell}))\Biggr\}
\end{multline}
where $\alpha_{\ell} = \ell/m$ and 
$\Delta A_{ij}^k(\alpha_{\ell}) = A_{ij}^k(\alpha_{\ell}) - A_{ij}^k(\alpha_{\ell-1})$.
Hence, the following is a numerical approximation of the~$w_k^c$:
\begin{equation}\label{e:igcam_ssum}
  w_k^c = \frac{1}{Z} \sum_{i,j} \left(\sum_{\ell=1}^m
    \Biggl\{\frac{\partial y^c(\alpha_{\ell})}{\partial A_{ij}^k}
    \times \Delta A_{ij}^k(\alpha_{\ell})\Biggr\}\right)
\end{equation}

As mentioned for Grad-CAM, 
optionally the integrated gradients can be
replaced with positive integrated gradients if we want to use only the units that
contribute positively to the output of the network:
\begin{equation}\label{e:igcam_ssum2}
  w_k^c = \frac{1}{Z} \sum_{i,j} 
    \text{ReLU}\left(\sum_{\ell=1}^m
    \Biggl\{\frac{\partial y^c(\alpha_{\ell})}{\partial A_{ij}^k}
    \times \Delta A_{ij}^k(\alpha_{\ell})\Biggr\}\right)
\end{equation}

\subsection{Summary of methods and theoretical contributions}
\label{summary}

We summarize our theoretical contributions by restating
the localization maps for each method and highlight their differences.
Unless otherwise stated, the localization maps will be
min-max normalized and resized to the shape of the original image.

\textbf{Grad-CAM.}
Equation (\ref{e:heatmap}) with weights explicitly written:
\begin{equation}
L^c_{\text{Grad-CAM}} = \\
\text{ReLU} \Biggl\{ \sum_k \Bigl( 
\underbrace{\frac{1}{Z}\sum_i \sum_j \frac{\partial y^c}{\partial A_{ij}^k}}_{w_k}
\Bigr) A^k \Biggr\}
\end{equation}

\textbf{Integrated Gradients.}
Equation (\ref{e:ig_to_layer_compact}):
\begin{multline}
L^c_{\text{Integrated Gradients}} \approx \\
 (A(1) - A(0))  \odot  \left( \frac{1}{m} \sum_{\ell=1}^m
 \frac{\partial y^c(\alpha_{\ell})}{\partial A} \right)
\end{multline}
where $\partial y^c(\alpha_{\ell})/\partial A$ represents the 3D tensor with elements
$\partial y^c(\alpha_{\ell})/\partial A_{ij}^k$,
and $\odot$ represents Hadamard (element-wise) product  \cite{horn2012}.
Note that the result is a 3D tensor
with the same shape as $A$, although a 2D localization map can be obtained
by adding the tensor feature maps across its channels.

\textbf{Integrated Grad-CAM.}
Equation (\ref{e:igradcam_approx}):
\begin{multline}
L^c_{\text{Integrated Grad-CAM}} \approx  \\
\frac{1}{m} \sum_{\ell=1}^m  \text{ReLU} \Biggl\{ \sum_k  \Bigl(
\underbrace{\sum_{i,j} \frac{\partial y^c(\alpha_{\ell})}{\partial A_{ij}^k}}_{w_k}
  \Bigr)
  \Delta_k(\alpha_{\ell}) \Biggr\}
\end{multline}
where $\Delta_k(\alpha_{\ell}) = A^k(\alpha_{\ell}) - A^k(0)$.
The method is equivalent to averaging Grad-CAM localization maps
obtained from a sequence of interpolated input images.

\textbf{RSI-Grad-CAM.}
Approximate integration of gradients with Riemann-Stieltjes sum (our method),
based on equation~(\ref{e:igcam_ssum}):
\begin{multline}
L^c_{\text{RSI-Grad-CAM}} \approx  \\
\text{ReLU} \Biggl\{ \sum_k 
\Biggl( 
\underbrace{\frac{1}{Z}\sum_{ij} \Bigl(
\sum_{\ell=1}^m \frac{\partial y^c(\alpha_{\ell})}{\partial A_{ij}^k}
\Delta A_{ij}^k(\alpha_{\ell})  \Bigr)}_{w_k} \Biggr) A^k \Biggr\}
\end{multline}
where $\Delta A_{ij}^k(\alpha_{\ell}) = A_{ij}^k(\alpha_{\ell}) - A_{ij}^k(\alpha_{\ell-1})$.

Gradients have been replaced with Riemann-Stieltjes integrated gradients
(approximated with Riemann-Stieltjes sums).

\subsection{Metrics}
\label{metrics}

We will compare the performance of Grad-CAM, Integrated Grad-CAM and our
RSI-Grad-CAM based on their numerical stability, and two quantitative 
evaluation approaches:

\begin{itemize}
    \item Quantitative Evaluations Without Ground Truth.
    \item Quantitative Evaluations With Ground Truth.
\end{itemize}

\subsubsection{Numerical stability}
\label{num_stability}

When computing mathematical operations we need to use a limited number 
of bits to express real numbers, which may lead to approximation errors.
Those errors can accumulate, leading to the failure of theoretically 
successful algorithms. One potential problem that may impact the algorithms
examined here is division by a near-zero number.

One of the steps in producing a heatmap for an attribution method is performing
min-max normalization, i.e., for each pixel 
with coordinates~$(i,j)$ we replace the heatmap pixel value $L(i,j)$ with 
\begin{equation}
L_{norm}(i,j) = \frac{L(i,j) - \min(L)}{\max(L) - \min(L)}
\end{equation}
so the normalized
pixel values range from 0 to 1. Since the pixel values of a heatmap are computed 
using gradients, and vanishing gradients are a common problem in some architectures,
the final result may involve zero or near zero pixel values in the raw (non-normalized) heatmap.

In order to avoid division by zero or near zero we will include a small term $\varepsilon$
in the bottom of the fraction used for normalization: 
\begin{equation}\label{e:normeps}
L_{norm}(i,j) = \frac{L(i,j) - \min(L)}{\max(L) - \min(L) + \varepsilon}
\end{equation}
This parameter $\varepsilon$ plays the role of an hyperparameter that needs to be
tweaked.  
If its value is too large it will distort the heatmap, 
if it is too low it won't be suitable to prevent underflow errors.

The range of values taken by
raw (non-normalized) heatmaps across an image dataset provides an indication of
the numerical stability of an attribution method.

\subsubsection{Quantitative Evaluations Without Ground Truth}
\label{eval_w_gt}

We use quantitative evaluations that do not require ground truth
as used in \cite{chattopadhyay2018}. The idea behind these kinds of evaluations is to measure
how much the prediction of the network changes when the original image is replaced by the
parts of the image highlighted by the heatmap produced by the attribution method being tested.

Given an image $I$ and a heatmap $L^c$ generated for this image 
for class $c$, we find an \emph{explanation map} $E^c = L^c \odot I$,
where $\odot$ represents the Hadamard (element-wise) product of $L^c$ and $I$. 
Since the
heatmap is a 2D array and the image has three color channels, the Hadamard product $E^c = L^c \odot I$
actually means the element-wise product of $L^c$ by each of the three channels of $I$, i.e., 
if the elements of $E^c$, $L^c$, and $I$ are respectively
$e^c_{ijk}$, $l^c_{ij}$, $p_{ijk}$, then $e^c_{ijk} = l^c_{ij} \cdot p_{ijk}$.

\begin{figure}[!htb]
\centering
\includegraphics[width=3.25in]{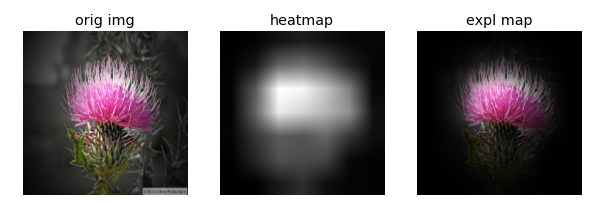}
\caption{Original image, heatmap, and explanation map.}
  \label{f:expl_map}
\end{figure}

{\figurename}\,\ref{f:expl_map} shows an example of image, heatmap, and resulting explanation map.

Feeding the network with an image $I$ we obtain an output $Y^c=$ predicted probability of class $c$.
If we feed the network with the explanation map $E^c$ we will obtain an output $O^c$.
For a good attribution method we expect $O^c$ to be close to 
the predicted probability $Y^c$.
Based on this idea, the following metrics
are defined:
\begin{align}
    &\text{Percentage Average Drop} 
    = \frac{100}{N} \sum_{i=1}^N \frac{\max(0,Y^{c_i}_i - O_i^{c_i})}{Y_i^{c_i}} \\
    &\text{Increase in Confidence} = \displaystyle{\frac{1}{N}\sum_{i=1}^N \mathbbm{1}(Y_i^{c_i} < O_i^{c_i})}
\end{align}
where $\mathbbm{1}$ is the indicator function with value 1 if the argument is true, and 0 if it is false, 
$i$ is an index running through the image dataset, 
and $c_i$ is the class predicted by the network when fed with the $i$th image.

Intuitively, the ``drop'' is the proportion of decrease 
of the network output when replacing the original image with the explanation map, and the Percentage Average Drop
is the average of the drop through the image dataset multiplied by 100 (lower is better).\footnote{In our
computations we replaced the definition of Percentage Average Drop by omitting the factor 100, 
so rather than a percentage we are using a proportion, 
yielding a value in the range 0--1 rather than 0--100.}
The Increase in Confidence is
the proportion of images for which the explanation map produces a network output larger than the network output produced by the original image (higher is better).

\subsubsection{Quantitative Evaluations With Ground Truth}
\label{eval_no_gt}

With an image dataset containing ground-truth bounding boxes, we can use  
metrics indicating in what extent the heatmaps overlapped the bounding boxes.  This was done in two ways:

\textbf{Pixel Energy}, defined as 
    $\frac{\sum L^c_{(i,j)\in bbox}}{\sum L^c_{(i,j)\in bbox} + \sum L^c_{(i,j)\notin bbox}}$,
    i.e., the sum of pixel intensities in the part of the heatmap inside the bounding box 
    divided by the total sum of intensities of the heatmap for the entire image
    (see energy-based pointing game in sec.~4.3 of \cite{wang2020scorecam}).
    When comparing two heatmaps 
    generated by the same input image, higher pixel energy is better. 
    Range goes from 0~to~1.
    
\begin{figure}[!htb]
\centering
\includegraphics[width=3.25in]{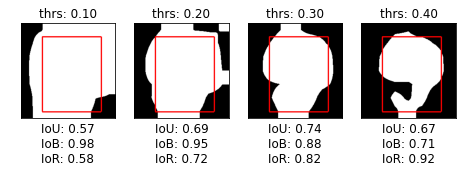}
\caption{Examples of Jaccard-based metrics for several threshold values.}
  \label{f:jaccard_ex}
\end{figure}

\textbf{Jaccard-based} measures: 
    for a given threshold, find the region $R$ occupied by the pixels of the
    heatmap whose intensities are above the threshold ({\figurename}\,\ref{f:jaccard_ex}). 
    Then determine how much the region overlaps with the 
    bounding box $B$ using Intersection over Union 
    $IoU = |R\cap B|/|R\cup B|$, Intersection over Bounding Box $IoB = |R\cap B|/|B|$, 
    and Intersection over Region $IoR = |R\cap B|/|R|$. 
    When comparing two heatmaps 
    generated by the same input image, higher Jaccard-based metrics are better.  Ranges go from 0~to~1.

\section{Implementation and Testing}

\begin{figure}[!htb]
\centering
\includegraphics[width=2.6in]{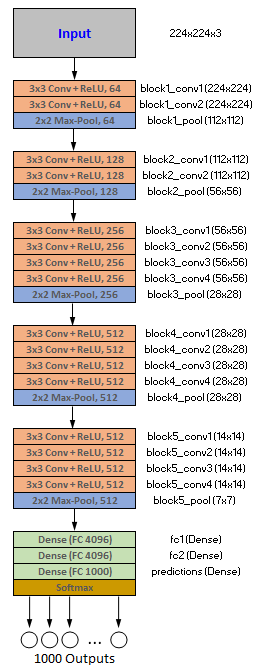}
\caption{VGG-19 network architecture with 1000 outputs 
corresponding to the 1000 classes used in training 
the network on a subset of the ImageNet dataset.}
\label{f:vgg19_arch}
\end{figure}

\subsection{Implementation}
We call our algorithm RSI-Grad-CAM (for ``Riemann-Stieltjes 
Integrated Gradient Class Activation Map''.
We implemented it using Keras
over Tensorflow on Google Colab.  We started with an implementation of
Grad-CAM, and then modified it to replace the gradients with integrated
gradients.  A set of interpolated images between a (black) baseline
and the final image are fed to the network, and activations and
the gradients at the chosen layer are collected.  For efficiency, the
images are not fed to the network one by one, but in batches, in a similar way
to what is done during network training.  In theory the whole set of
images could be fed as a single batch, but this tended to exhaust
GPU resources, so we used limited size batches (32 images per batch).

The computation of the weights $w_k^c$ and the final linear
combination of feature maps $\sum_k w_k^c A^k$ are straightforward.
The application of a ReLU at the end allows to select only the units
that contribute positively to the score of the selected class.

In our implementation, when computing the weights,
we also selected only units in which activations,
integrated gradients, and activation total increments
$(A_{ij}^k(m) - A_{ij}^k(0))$ are all positive.\footnote{This 
part of the implementation is optional, in actual applications we
recommend to experiment with different rules of unit selection
to determine which one works best.}
This (inspired on the
``guided Grad-CAM'' approach of \cite{selvaraju2017grad}) allows the
algorithm to ignore extraneous elements that do not contribute to the
chosen class score, e.g. if an image contains a `dog' and a `cat', and
we are interested in locating only the dog, the area of the image
containing the cat is expected to produce:
\begin{itemize}
\item negative integrated gradients for the network output
  corresponding to `dog', and
\item negative activations and activation increments in the feature
  maps more strongly linked to the `dog' output.
\end{itemize}

As a consequence we expect that ignoring those units will produce
sharper heatmaps better focused in locating the elements of the image
related to the output of the chosen class.
For the comparison to be fair we also used the Grad-CAM
version with positive gradients in which weights are computed using
equation~(\ref{e:gradcam_weights2}).

After a heatmap has been produced at the layer level, it is upsampled to
the original size and overlaid to highlight the elements of the input
image that most contribute to the output corresponding to the desired
class.

For the baseline we acknowledge that various choices are possible
(black image, random pixels image, blurred image, etc.).  In the present
work we follow the choice suggested by the 
authors of the Integrated Gradients algorithm
in~\cite{sundararajan2017ig}
for object recognition networks, i.e., a black image.

\subsection{Preliminary Testing}

We compare our RSI-Grad-CAM to Grad-CAM and Integrated Grad-CAM
on a VGG-19 network \cite{simonayan2015}.

The VGG-19 model consists of five blocks, 
each containing a few convolution layers followed 
by a max-pooling layer. For convolution operation,
VGG-19 uses kernels of size $3\times 3$. 
The layers on each of its five blocks have
64, 128, 256, 512 and 512 channels respectively,
as shown in {\figurename}\,\ref{f:vgg19_arch}.
The final layer has one thousand units and a softmax activation function, so its 
final outputs can be interpreted as a vector of one thousand probabilities, one per class.

\begin{figure}[!htb]
\centering
\includegraphics[width=2.8in]{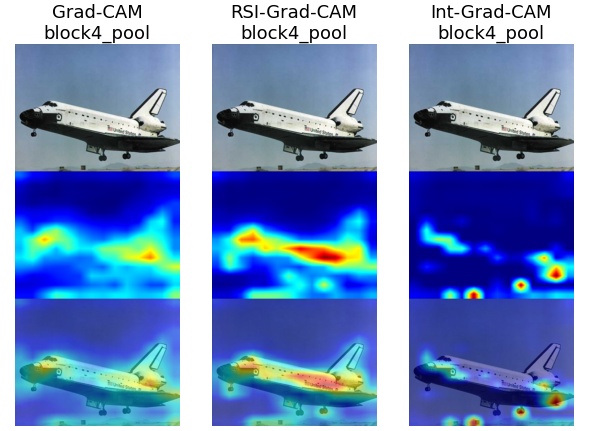}
\caption{First row: original image with 100\% prediction probability; 
second row: heatmaps produced by Grad-CAM, RSI-Grad-CAM, and Integrated Grad-CAM
respectively for the image; third row: overlaid heatmaps.}
\label{f:shuttle}
\end{figure}

\begin{figure*}[!htb]
\centering
\includegraphics[height=2.3in]{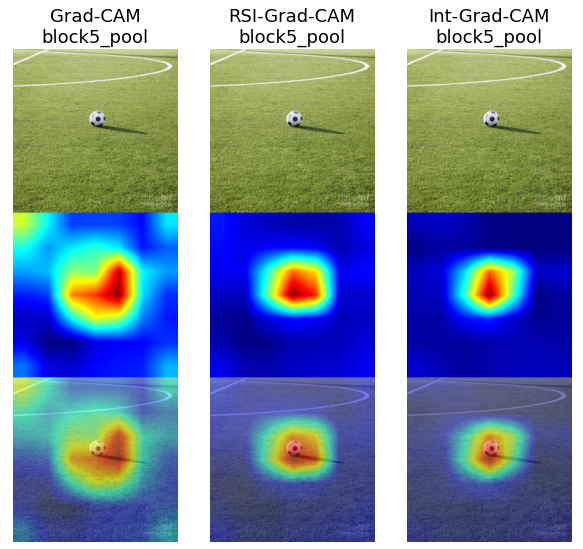}
\hskip 10pt
\includegraphics[height=2.3in]{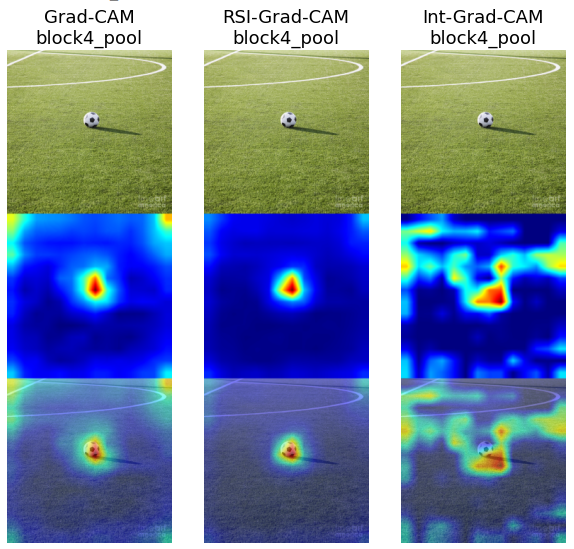}
\caption{Heatmaps produced for a soccer ball on a paying field
at the last block (left), and second to last (right).}
\label{f:soccer_ball}
\end{figure*}

\begin{figure*}[!htb]
\centering
\includegraphics[height=1.8in]{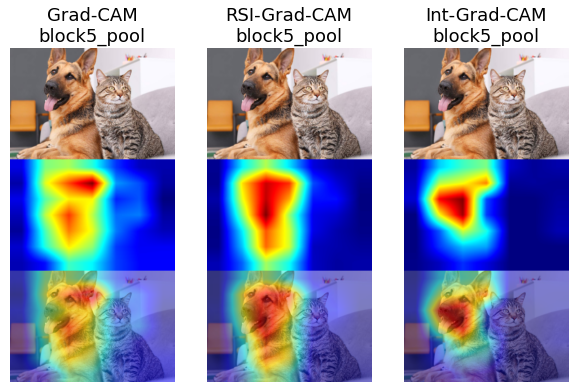}
\hskip 20pt
\includegraphics[height=1.8in]{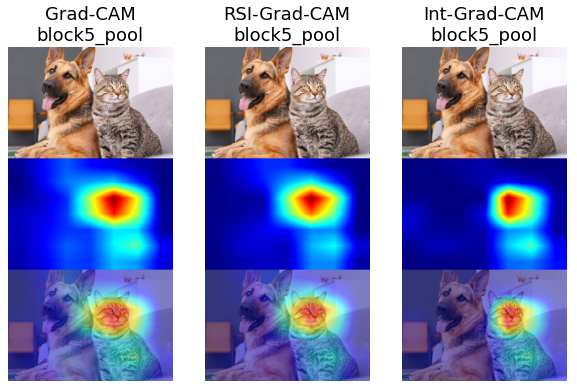}
\caption{Images with a German Shepherd and a cat. Heatmaps for the
  German Shepherd (top) and the cat (bottom).  The score 
  assigned to the cat is very low (0.22\%), hence far from saturation, and in
  this case we do not expect Grad-CAM and RSI-Grad-CAM to 
  yield very different heatmaps.  The difference is more noticeable for the dog.
  Integrated Grad-CAM does a good job on this layer, separating correctly dog and cat.}
  \label{f:german_shepherd_and_cat}
\end{figure*}

\begin{figure*}[!htb]
\centering
\includegraphics[height=2in]{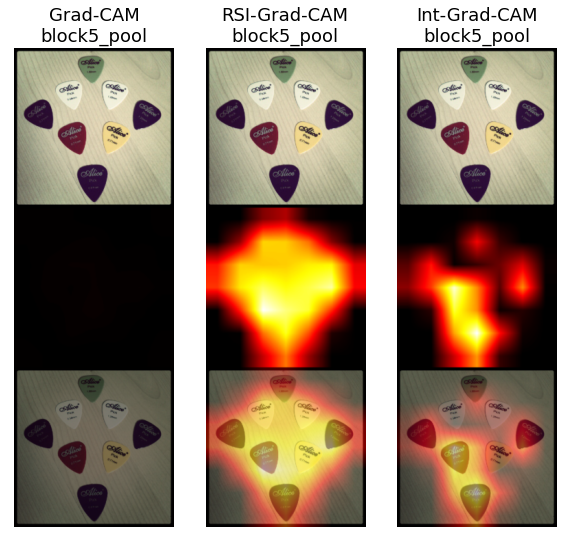}
\hskip 20pt
\includegraphics[height=2in]{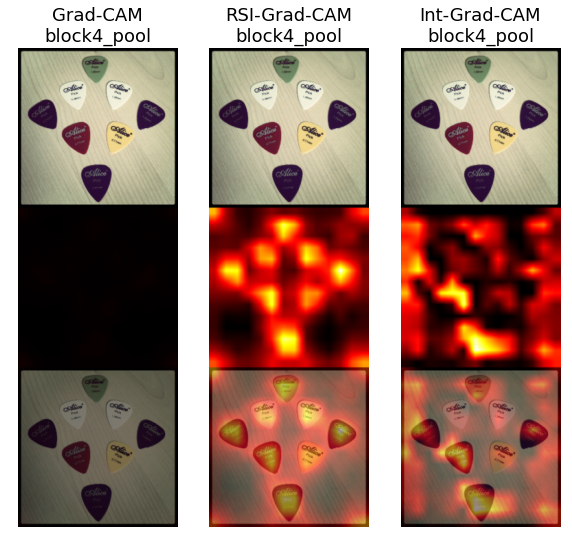}
\caption{Images of a set of guitar picks produced at the last convolutional block (left)
and the next to the last block (right). Note how our RSI-Grad-CAM at the last block highlights the general area occupied by the set of picks, 
and it focuses on individual picks at the next to the last block.
Grad-CAM produces dark heatmaps, and Integrated Grad-CAM fails to produce
a useful heatmap at the last layer of block 4 (block4\_pool).}
  \label{f:picks}
\end{figure*}

All three attribution methods tested here are applied to the final pooling layer 
of each convolutional block. When we say that an attribution method is 
applied to the $n$th layer of the network that means ``to the
final polling layer of the $n$th block.''

{\figurename}~\ref{f:shuttle}--\ref{f:picks}
show examples of scenarios in which our proposed approach
RSI-GradCAM produces better visualization maps than Grad-CAM and Integrated
Grad-CAM.

{\figurename}\,\ref{f:shuttle} 
shows the heatmaps of an image for which
the probability of the prediction by
the network is 100\%, so the output clearly becomes saturated.
As expected, the output of the usual Grad-CAM is tenuous, while
RSI-Grad-CAM yields a more visible heatmap. Integrated Grad-CAM produces
a blurry heatmap a this level.

{\figurename}\,\ref{f:soccer_ball} shows the heatmaps for a soccer
ball at the last two layers. The network output for this image
was 85.6\%, far from saturation.  Compared to Grad-CAM, 
at the last and second to last blocks of the network,
Integrated Grad-CAM and RSI-Grad-CAM
produce sharper heatmaps, better focused on the soccer~ball.

{\figurename}\,\ref{f:german_shepherd_and_cat}
illustrate the ability to locate two
different elements in the same image.  The network detects the German
Shepherd and assigns it a 67.79\% score, and the cat with 0.6\% score.
For the German Shepherd the heatmap produced by RSI-Grad-CAM is 
brighter and sharper than the one produced by Grad-CAM. The
difference is less noticeable for the cat, for which the network
output is far from saturation.  The fact that the heatmaps highlight
mainly the animal's head is an indication that its body plays a lesser
role in identifying the species. Integrated Grad-CAM does a good job 
locating both animals.

In general we observe that, when the network outputs are far from saturation,
the heatmaps produced by our algorithm are similar to that of the the
original Grad-CAM, the difference is more noticeable when the outputs
are almost saturated.  We also observe that the heatmaps produced by
our algorithm look sharper and better focused in the area of the
image that contains the desired element or feature.  

{\figurename}\,\ref{f:picks} shows heatmaps for a set of guitar picks. 
The probability predicted by the network was 100\%, and the saturation of the output
was so high that the heatmap produced by Grad-CAM was too tenuous to be visible. Heatmaps produced
by our RSI-Grad-CAM highlights the whole set of picks at the last convolutional block (block5\_pool layer),
and individual picks at the next to last block (block4\_pool layer).  
This example illustrates the fact that heatmaps produced at
different layers highlight different aspects of the image: global (such as the general area 
occupied by the object of interest) when close to the output, and local details
(such as individual parts) when produced at intermediate layers.

\subsection{Quantitative Evaluations}

The examples shown in the previous section are illustrative.
Here we use the quantitative metrics 
introduced in section~\ref{metrics}
to evaluate our attribution technique.
For that purpose, we use a common image classification network working 
on a fairly large dataset with a variety of images
of objects and natural elements, as detailed below.

\subsubsection{Dataset and Model}
\label{dataset-model}

\begin{figure}[!htb]
\centering
\includegraphics[width=1.25in]{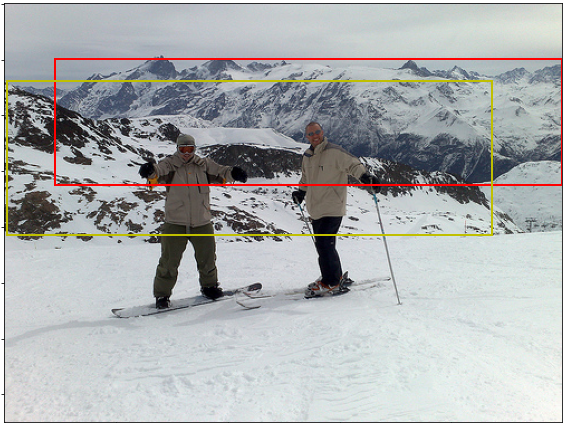}
\hskip 20pt
\includegraphics[width=1.25in]{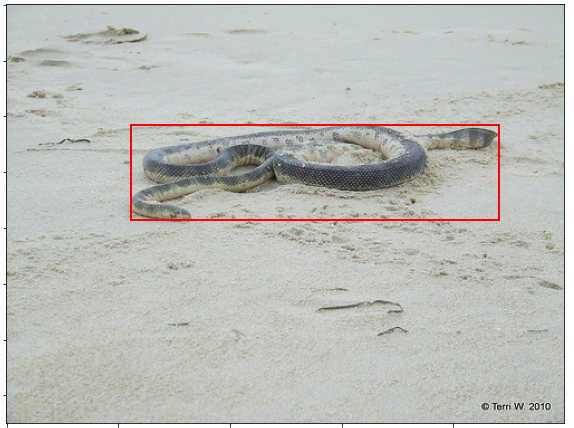}
\caption{Two images from the ILSVRC2012 dataset. Left: image of Alp mountains with two bounding boxes.
Right: image of a sea snake, with one bounding box. The bounding boxes are shown for illustration
purposes only, the network was fed with images that didn't display bounding boxes.}
  \label{f:allp_and_sea_snake}
\end{figure}

We used the VGG-19 network pretrained on ImageNet \cite{simonayan2015},
with input shape $224\times 224\times 3$,
and performed experiments on a subset
of the validation dataset used for the
ImageNet Large Scale Visual Recognition Challenge 2012 (ILSVRC2012) \cite{russakovsky2015}.
The ILSVRC2012 dataset is a subset of ImageNet
with 50,000 images from 1,000 categories, annotated with labels
and rectangular bounding boxes obtained using the 
Amazon Mechanical Turk \cite{sorokin2008}. {\figurename}\,\ref{f:allp_and_sea_snake} shows two randomly selected images  with their bounding boxes.

In all the tests we picked a convolutional block and computed heatmaps generated
by each of the attribution methods at the maxpooling layer of the block.

The subset of images was chosen so that:
\begin{enumerate}
\item[1.] The network predicted the right class, to make sure that we are
evaluating the attribution technique rather than the network performance.
\item[2.] The image contained only one bounding box, to avoid images with more than one category,
or multiples occurrences of the same category.
\item[3.] The bounding box occupied less that 50\% of the image area, because an object that
occupies most of the image would be too easy to locate and would produce unreliably metrics.
\end{enumerate}
The final dataset used contained a total of 12,525 images.

In the next section we compare the performance of Grad-CAM, RSI-Grad-CAM and Integrated Grad-CAM 
in three aspects: numerical stability, quantitative evaluations without ground truth, and 
quantitative evaluations with ground truth.

\begin{figure}[!htb]
\centering
\includegraphics[width=3.25in]{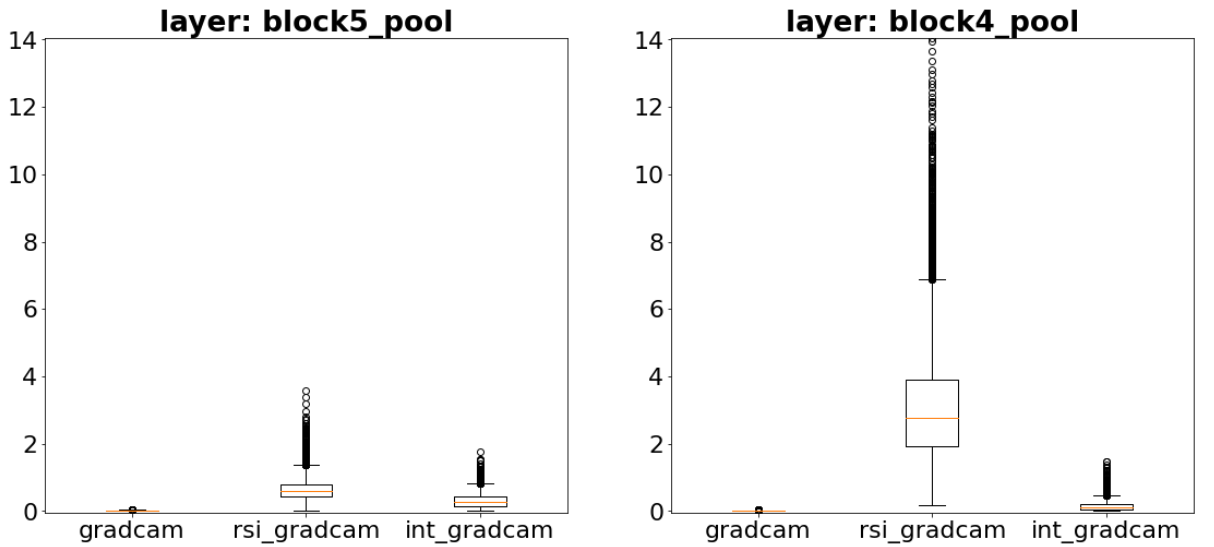}
\caption{Boxplot of mean pixel values of heatmaps before normalization
at the last two convolutional blocks.}
  \label{f:heatmap_int_raw}
\end{figure}

\begin{figure}[!htb]
\centering
\includegraphics[width=3.25in]{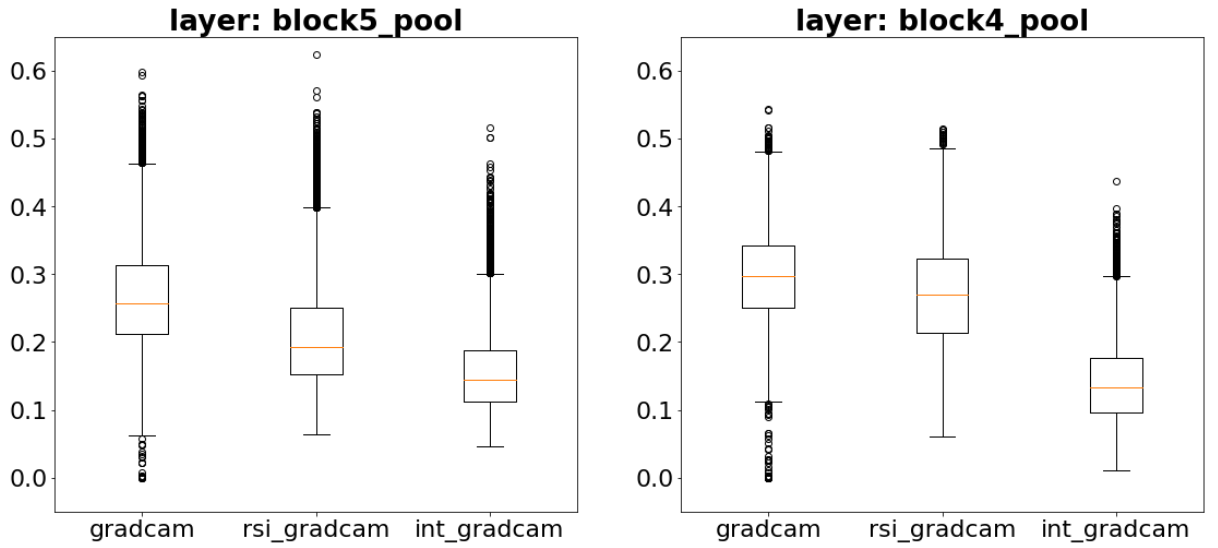}
\caption{Boxplot of mean pixel values of heatmap after normalization
at the last two convolutional blocks.}
\label{f:heatmap_int_normal}
\end{figure}

\begin{figure*}[!htb]
\centering
\includegraphics[height=2.5in]{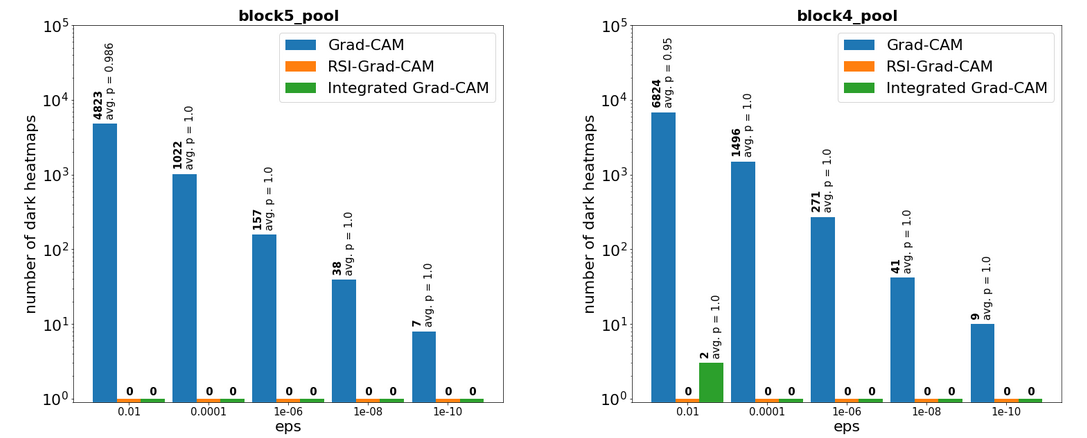}
\caption{Number of dark heatmaps produced for the attribution
methods applied to the last and next to the last blocks.}
  \label{f:dark-heatmaps-barplots}
\end{figure*}

\begin{figure*}[!htb]
\centering
\includegraphics[height=1.75in]{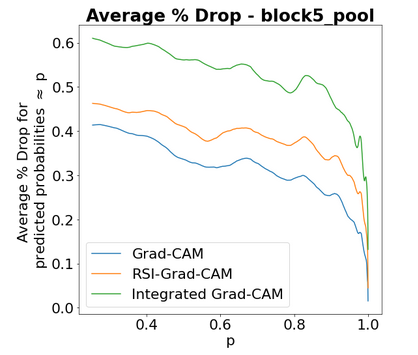}
\hspace{30pt}
\includegraphics[height=1.75in]{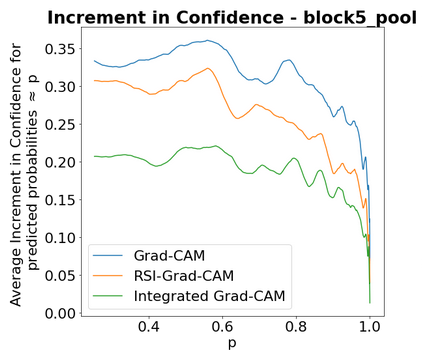}
\caption{Average \% Drop and Increment in Confidence at the last convolutional block.}
  \label{f:increment-in-confidence-block5}
\end{figure*}

\begin{figure*}[!htb]
\centering
\includegraphics[height=1.75in]{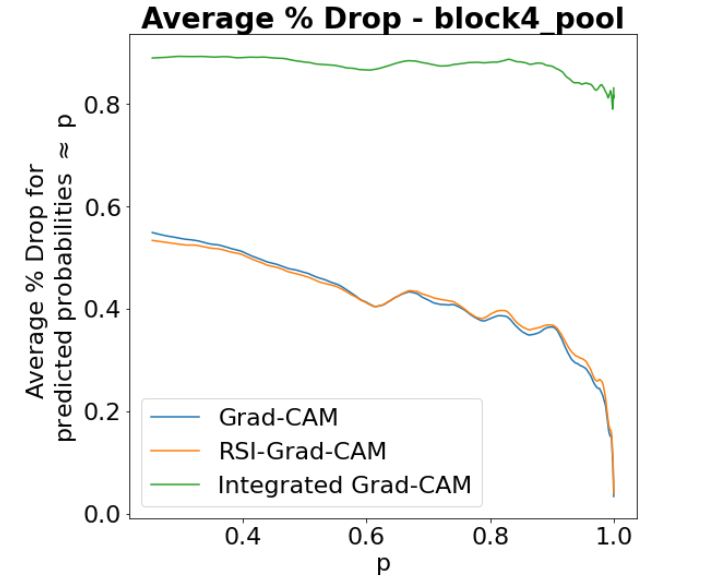}
\hspace{30pt}
\includegraphics[height=1.75in]{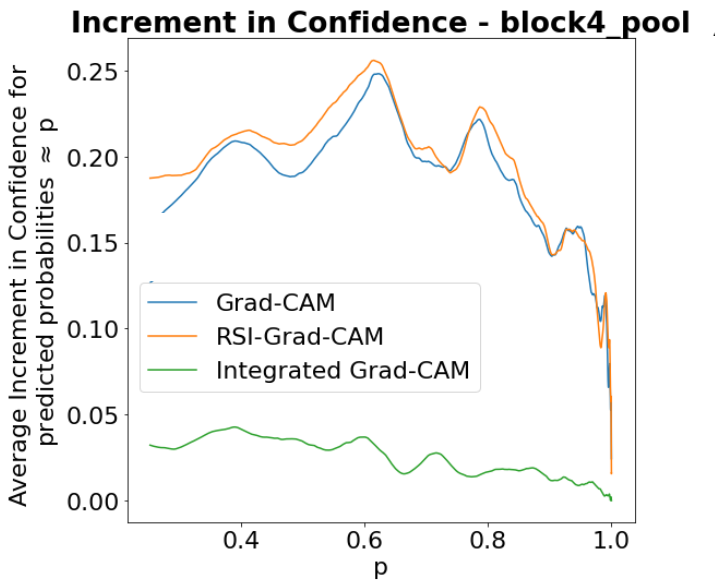}
\caption{Average \% Drop and Increment in Confidence at the next to the last convolutional block.}
  \label{f:increment-in-confidence-block4}
\end{figure*}

\begin{figure*}[!htb]
\centering
\includegraphics[width=3in]{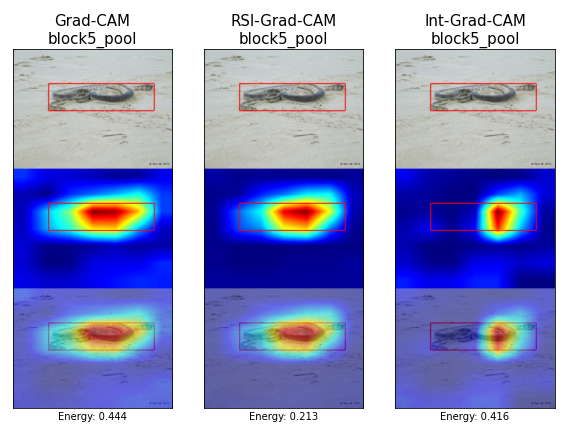}
\includegraphics[width=3in]{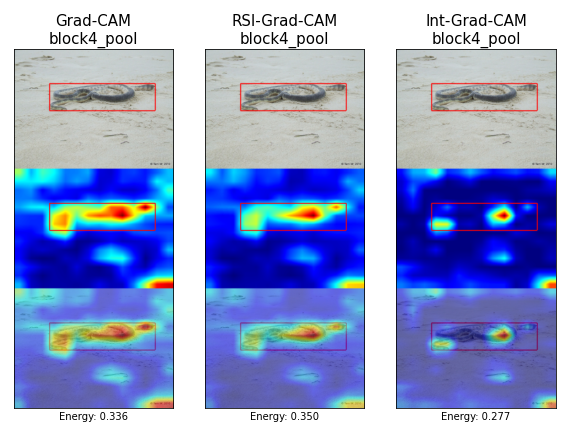}
\caption{Energy at the last two convolutional blocks for each attribution technique.
The colormap used to display the heatmap has been changed in this image
to make the overlapping between heatmap and bounding box more visible.}
  \label{f:energy_ex}
\end{figure*}

\begin{figure*}[!htb]
\centering
\includegraphics[height=1.75in]{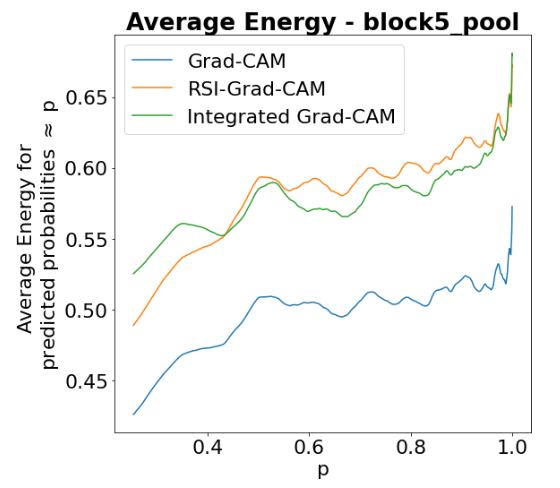}
\hspace{30pt}
\includegraphics[height=1.75in]{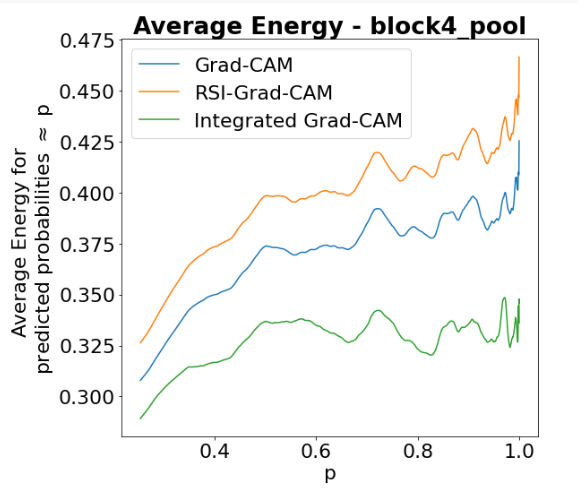}
\caption{Average energy at the last and next to the last convolutional blocks}
  \label{f:average-energy}
\end{figure*}

\subsubsection{Numerical Stability}

As explained in section~\ref{num_stability}, low pixel values of
heatmaps may introduce problems because of numerical instability.
{\figurename}\,\ref{f:heatmap_int_raw} shows that Grad-CAM is particularly
sensitive to this phenomenon. For each image we produce the corresponding (non-normalized)
heatmap, and compute the mean value of its pixel values. The boxplots in 
{\figurename}\,\ref{f:heatmap_int_raw} show the distribution of mean pixel values of
heatmaps produced for our database and model at the pooling layers of the last two convolutional
blocks (block4\_pool and block5\_pool respectively) for each of the attribution
methods tested.

We can see that the pixel values of the heatmap produced by 
Grad-CAM are very close to zero. This can cause the denominator of the fraction
used for normalization of the heatmap to become zero or near zero,
producing floating point underflow and unreliable results.
In order to avoid division by zero or near zero we include a small term $\varepsilon$
in the bottom of the fraction used for normalization of a heatmap $L$, 
as shown in equation (\ref{e:normeps}).
In our tests
we used $\varepsilon = 10^{-8}$.
{\figurename}\,\ref{f:heatmap_int_normal} 
shows the result
of normalizing using this correction term.

We note that even after normalizing
Grad-CAM is still producing some heatmaps with 
a mean value near zero.
To avoid a blank heatmap, 
it is possible to reduce the value of $\varepsilon$ even more,
although it is hard to tell exactly by how much without some trial and error.

We will define a \emph{dark heatmap} as one
in which its maximum pixel intensity is less than half the maximum
possible range of pixel values, e.g. less than $0.5$ in a scale from 0~to~1:
$(L_{max} - L_{min})/(L_{max} - L_{min} + \varepsilon) < 0.5$.
This happens precisely for $L_{max} - L_{min} < \varepsilon$
in equation (\ref{e:normeps}).
Note that the $0.5$ threshold value used in the definition of ``dark heatmap''
could be changed to some other value inside the interval $(0,1)$
without essentially changing the discussion below.
E.g., if we replace it with
$0.25$, the new ``dark'' images would look even darker, and there will be fewer of them,
but that would not change the argument presented below in a meaningful way,
it would just change
the $L_{max} - L_{min} < \varepsilon$ condition to the one obtained from
$(L_{max} - L_{min})/(L_{max} - L_{min} + \varepsilon) < 0.25$, i.e.
$L_{max} - L_{min} < \varepsilon/3$, so the analysis presented below would
be the same just replacing $\varepsilon$ with $\varepsilon/3$.
Other than that the barplots in 
{\figurename}\,\ref{f:dark-heatmaps-barplots} would look
the same, just with different numerical labels in the 'eps' axis,
and the comparison among the different attribution methods regarding the 
risk of producing ``dark heatmaps'' would be the same too.

Now we can count the number of
dark heatmaps across the image dataset produced by a given attribution
method applied to a given layer. 
The barplots in
{\figurename}\,\ref{f:dark-heatmaps-barplots}
show the result of using various
values of $\varepsilon$  (called `eps' in the figure). 
Note that the plots are drawn using  logarithmic scales. 
Here `avg~p' is the average value of the network output 
(probability of the input image belonging to the corresponding class) 
for all the images for which the attribution method
produces a dark heatmap. It is included 
to show how it correlates with the number of dark heatmaps.
We see that, for Grad-CAM dark heatmaps are produced when the 
network output gets close to $1$, the point at which 
it gets saturated and the vanishing gradient problem arise.

Another approach is to use Grad-CAM on the network with the
final softmax layer removed to avoid the effects of vanishing gradients caused by
saturation of the network output.
In any case this shows the fragility of Grad-CAM, which may require to tweak parameters and network
architecture to make it work. Integrated Grad-CAM and our technique RSI-Grad-CAM
are not affected by these problems.

\begin{figure*}[!htb]
\centering
\includegraphics[height=2in]{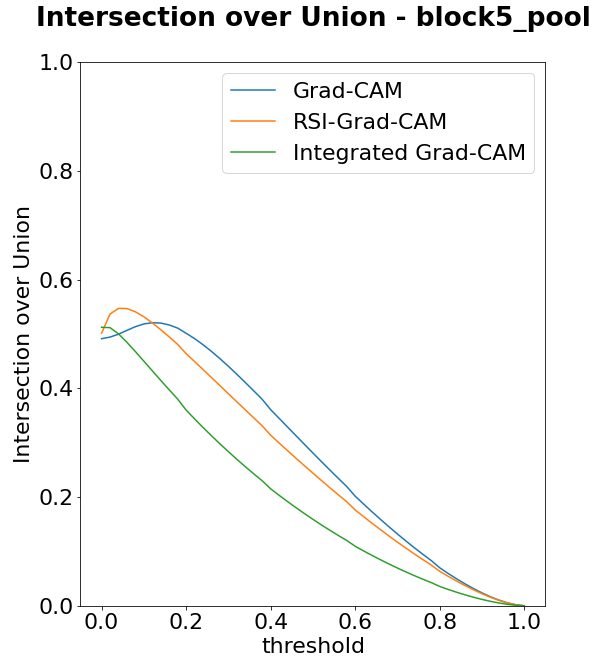}
\hspace{10pt}
\includegraphics[height=2in]{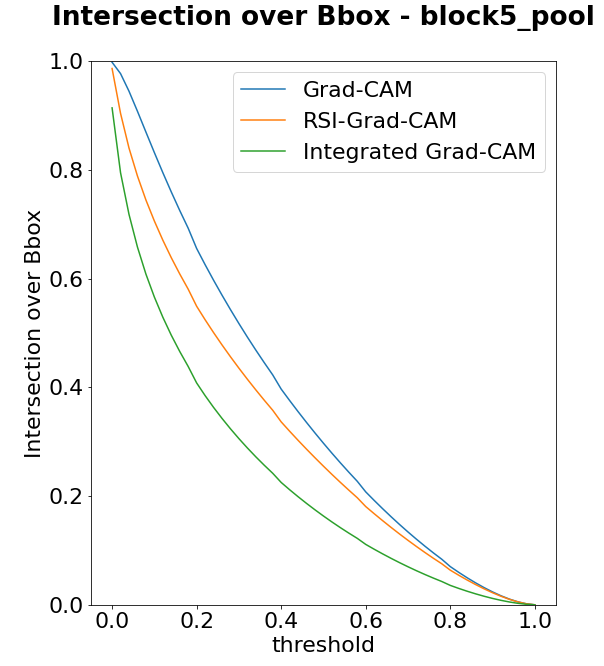}
\hspace{10pt}
\includegraphics[height=2in]{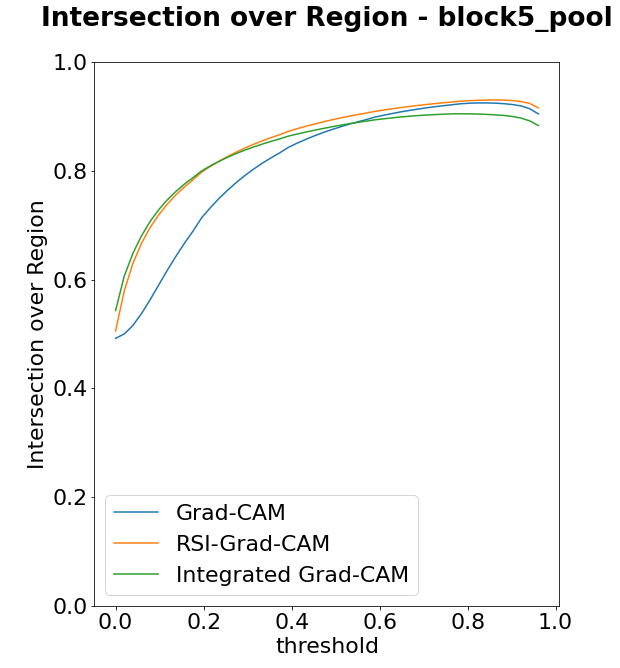}
\caption{Intersection over Union, Intersection over Box, 
and Intersection over Region at the last convolutional block.}
  \label{f:jac-block5}
\end{figure*}

\begin{figure*}[!htb]
\centering
\includegraphics[height=2in]{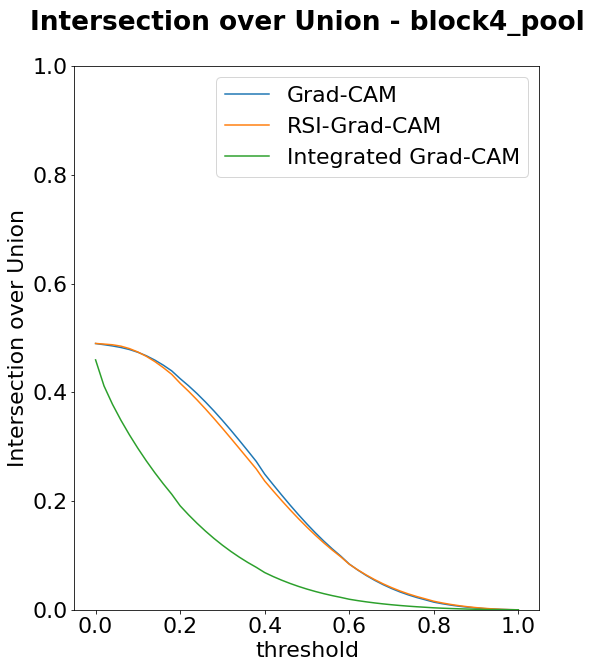}
\hspace{10pt}
\includegraphics[height=2in]{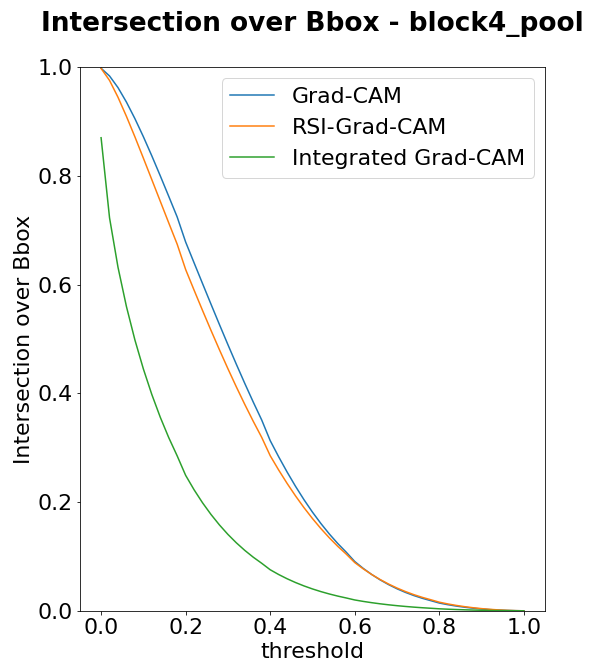}
\hspace{10pt}
\includegraphics[height=2in]{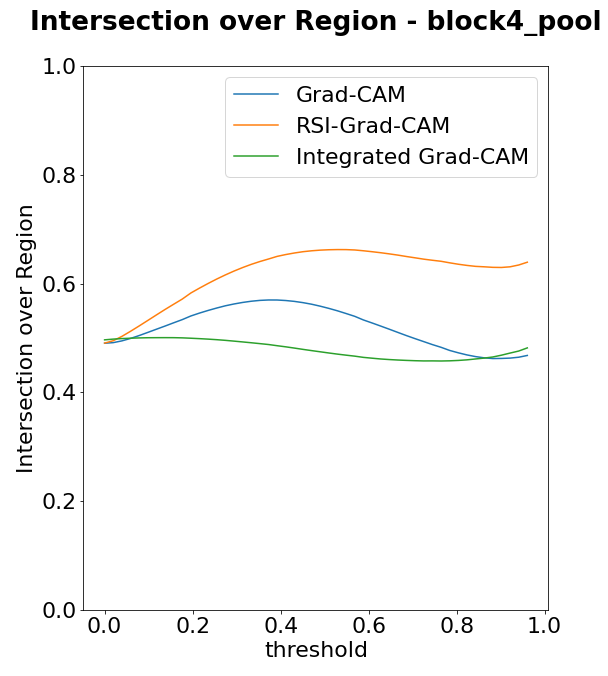}
\caption{Intersection over Union, Intersection over Box, 
and Intersection over Region at the next to the last convolutional block.}
  \label{f:jac-block4}
\end{figure*}

\subsubsection{Quantitative Evaluations Without Ground Truth (Results).}

Now we look at the results of applying Average Drop and
Increment in Confidence, introduced in section~\ref{eval_w_gt}.

In order to determine if these metrics depend on the predicted probability, the graphics shown
in {\figurename}\,\ref{f:increment-in-confidence-block5}~and~\ref{f:increment-in-confidence-block4}
were obtained by sorting the images by predicted probability, and computing averages across a fix 
length rolling window. In our experiments the window 
had a width of 1000 samples, so 
each point in the graph represents
the average metric obtained for 
1000 consecutive samples
(the $p$ coordinate in the graph is also the average probability of the 1000 elements contained in the sliding window).
We found that the performance
of Grad-CAM was the best at the last convolutional block, but our method RSI-Grad-CAM
did slightly better than Grad-CAM, while the performance of Integrated Grad-CAM got worse
when computed at blocks below the last one.

\subsubsection{Quantitative Evaluations With Ground Truth (Results).}

Here we look at the results of applying Pixel Energy and
Jaccard-based measures, introduced in section~\ref{eval_no_gt}.
{\figurename}\,\ref{f:energy_ex} provides an example of how the 
pixel intensity is distributed in relation to the bounding box.
The results for the whole dataset (as a function of the probability $p$
predicted by the network)
are shown in {\figurename}\,\ref{f:average-energy}. We observe that
our method RSI-Grad-CAM performs better than Grad-CAM at the 
last and next to the last convolutional blocks of the network. 
Compared to Integrated Grad-Cam, RSI-Grad-CAM produces similar results
at the last convolutional block, but again it performs better at the
next to the last block.

On the other hand, the Jaccard-based measures depend on the threshold
chosen, as illustrated in {\figurename}\,\ref{f:jaccard_ex}.

{\figurename}~\ref{f:jac-block5} and \ref{f:jac-block4}
show that, compared to the other methods, 
our RSI-Grad-CAM performs better for Intersection over Union 
at the next to the last block, indicating that a larger proportion of the heatmap
tends to lie inside the ground-truth bounding box at that level.
This is consistent with the effect shown in {\figurename}\,\ref{f:picks}, where the heatmap produced by 
RSI-Grad-CAM highlights internal structures of the object of interest
(in this case individual picks in a set of picks). This makes our attribution method ideal
in picking up middle level features of an image.

\section{Conclusions}

We have examined three attribution techniques intended to provide
explanations about how CNNs make their predictions, and proposed a new
method that better implements the goals of those techniques.

Grad-CAM uses gradients of the network output for a given class
computed at an arbitrary convolutional layer.  Those gradients are used
to determine the relative contribution of each feature map in that
layer to produce a heatmap highlighting the regions of the network
input that contribute to the network output.  While this technique
works relatively well in many situations, its performance suffers when
any of the network layers, particularly its output layer, is near saturation level.
Integrated Gradients overcomes the problem caused by the network
output saturation by integrating the gradients of the network outputs
with respect to the inputs of the network along a set of outputs obtained
by interpolation from a baseline to the desired input. However, it may
miss features captured at hidden layers of the network.

Integrated Grad-CAM offers a solution based on integrating saliency maps.
We include it here for comparison to our method, RSI-Grad-CAM.
Unlike Integrated Grad-CAM, our RSI-Grad-CAM method replaces
the gradients used by Grad-CAM
with gradients integrated using the activations of the units of the
internal layer as integrators.  
Compared to Integrated Grad-CAM,
our method have comparable performances only
when used at the last layer of the network, but the performance
of Integrated Grad-CAM degrades quickly when used at hidden layers
below the last one, and then our method performs better at those layers.

Compared to Grad-CAM we observe that the results
of applying our RSI-Grad-CAM method yields better results when applied
to images in which the network outputs are near saturation,
has better numerical stability, 
and it is better suited to detect small details within the region
of interest when used at layers right below the last one.


\section{Future Work}

Any method based on line integrals depends on the integration path used.
In our algorithm we feed the network using a set of images linearly
interpolated between a baseline and the desired input.  
A possible area of research would be to explore alternate
integration paths.

On the other hand, there is a degree of arbitrariness in the choice 
of the baseline (a blank image in our case). Ideally the baseline should be 
an input that produces equal outputs for all classes. However it is unlikely that
only one output has such property, so additional conditions on the baseline
may need to be imposed depending on heuristic arguments (such as darker
baselines being preferred to bright ones as indicative of ``no features present'')
and practical considerations such as final performance.

Another line of work would be to replace Grad-CAM with our RSI-Grad-CAM
algorithm in existing works, such as the method proposed in 
\cite{chen2020embed} for use in embedding networks.


\end{document}